\documentclass[11pt]{article}
\usepackage[utf8]{inputenc}
\usepackage[font=small,format=plain,labelfont=bf,up,textfont=normal,up]{caption}
\usepackage{natbib}
\usepackage{graphicx}
\usepackage[hang, flushmargin, bottom]{footmisc}
\usepackage[margin=21mm]{geometry}
\usepackage{url}
\usepackage{setspace}
\usepackage{adjustbox}
\usepackage{amsmath}
\usepackage{courier}
\usepackage{threeparttable}
\usepackage{booktabs}
\usepackage{subfig}
\usepackage{multirow}
\usepackage{csquotes}  
\usepackage{hyperref}

\title{Bias in LLMs as Annotators: The Effect of Party Cues on Labelling Decision by Large Language Models}
\author{Sebasti\'an Vallejo Vera, University of Western Ontario, sebastian.vallejo@uwo.ca\thanks{Corresponding author. We would like to thank Laurenz Ennser-Jedenastik and Thomas M. Meyer from the University of Vienna for kindly sharing their data and code with us.} \\
Hunter Driggers, University of Western Ontario, mdrigger@uwo.ca. \vspace{0.3cm}} %
\date{}

\begin{document} 

\maketitle

\thispagestyle{empty}

\begin{abstract}
\noindent Human coders are biased. We test similar biases in Large Language Models (LLMs) as annotators. By replicating an experiment run by \cite{ennser2018impact}, we find evidence that LLMs use political information, and specifically party cues, to judge political statements. Not only do LLMs use relevant information to contextualize whether a statement is positive, negative, or neutral based on the party cue, they also reflect the biases of the human-generated data upon which they have been trained. We also find that unlike humans, who are only biased when faced with statements from extreme parties, LLMs exhibit significant bias even when prompted with statements from center-left and center-right parties. The implications of our findings are discussed in the conclusion.
\end{abstract}

\pagebreak
\setcounter{page}{1}
\setstretch{2}

\section{Introduction}

The increasing sophistication of large language models (LLMs) has allowed for their more prominent presence in political science research. One particular area gathering significant attention in the field is the use of LLMs as annotators. Research has shown promising results, with LLMs often outperforming human coders \citep{gilardi2023chatgpt} and providing comparable accuracy when labelling political text, across multiple languages \citep{heseltine2024large}. While researchers have evaluated the performance of LLMs as annotators across different domains, there still little information on how the known biases of LLMs \citep[see][]{gallegos2024bias} can affect their performance. 

For human annotators, studies show that political cues, such as party, have an effect on their coding decisions \citep{laver2000estimating,benoit2016crowd,ennser2018impact}. In this research note, we test whether this is also true for LLMs. We replicate the experimental design from \cite{ennser2018impact}, who tested coder bias by having humans annotate policy statements on immigration connected with party cues. We use the same treatment (i.e., the same policy statement with different party cues) to evaluate how two LLM families, OpenAI's ChatGPT and Meta's LLaMa, determine the sentiments behind policy statements (i.e., positive, negative, or neutral). The results show important differences in internal consistency across LLM families, low agreement between LLM and human coders and, most importantly, significant discrepancies conditional on treatment. In our conclusion, we discuss the implications of our results with respect to the use of LLMs as political text annotators.

\section{Bias in Annotation}

Numerous political behavior studies have shown that individuals' perceptions can be biased based on characteristics that are relevant to political preference formation: gender, education, racial identity, and partisanship. Since annotators are not immune to their political contexts, similar biases have been observed in coding tasks. In a wide-ranging meta-analysis, \cite{webb2023conservatives} show various sources of annotator bias, including partisanship and gender, when completing subjective and objective coding tasks.\footnote{\cite{webb2023conservatives} also find that there is low generalizability of the results. For example, gender biases found in U.S. annotators were not found in Dutch annotators.} Other research finds that partisanship can affect reactions of disgust \citep{ahn2014nonpolitical} and responses on opinion surveys \citep{schaffner2018misinformation,bullock2019partisan}. In the study that we replicate in this paper, \cite{ennser2018impact} show that coders use heuristics from party labels when judging political statements.

Research on LLMs has also explored biases and incongruities on their responses. Despite the promises and fanfare from LLM developers, studies have shown multiple sources of errors \citep{hicks2024chatgpt} as well as political bias in their output \citep{urman2023silence,rotaru2024artificial}. \cite{motoki2024more}, for example, find that LLMs tend to align more with left-of-center viewpoints, a result similar to the one obtained by \cite{rozado2024political} when probing LLMs with political orientation tests. More broadly, studies on LLMs have consistently found biases based on contextual and cultural factors \citep{gallegos2024bias}, leading to the misrepresentation of certain social groups \citep{yang2022unified}, gender stereotyping \citep{dong2024disclosure}, and the reinforcement of normativity \citep{bender2021dangers}. 

Despite the known limitation of LLMs, most research on LLMs as annotators has focused solely on accuracy, comparing their performance with that of human annotators. Relevant political science research has shown that LLMs outperform human annotators, at a fraction of the price, with minimal effects on downstream performance \citep[see][]{braylan2022measuring,heseltine2024large,gilardi2023chatgpt}. While the results from these studies provide promising avenues for the application of LLM as coders, less attention has been given to the effect of known LLM biases on performance. 

In this research note, we test the possible biases in LLM annotation resulting from political cues. We argue that LLMs will use political contextual information to evaluate statements and produce responses.\footnote{This is similar to heuristic processing in human annotator when using political party cues to evaluate statements \citep{ennser2018impact}.} As a simplified explanation, Transformer-based LLM assign representations (i.e., embeddings) to tokens by relating their occurrence in conjunction with other tokens in the text. To do this on a large scale, LLMs are fed enormous amounts of text from various sources.\footnote{For a more detailed explanation on the encoding-decoding infrastructure of Transformer-based model, see \cite{c2024bert}.} These corpora are not created in a vacuum, instead reflecting social realities. Thus, if certain parties (e.g., far-right parties) are often mentioned in certain contexts (e.g., discriminatory practices), LLMs are more likely to associate these parties with those contexts. When LLMs annotate text, they will use high-information tokens (i.e., political parties) to guide their responses, just as human use parties cues as a heuristic to evaluate statements \citep{ennser2018impact}. In the reminder of the paper, we evaluate to what degree party cues affect the output of LLMs when used as annotators.  

\section{Replication Setup}

To test possible biases in LLMs as annotators, we adapt \cite{ennser2018impact}'s experiment on party cues and human annotators. In their study, \cite{ennser2018impact} have ten coders classify 200 policy statements on immigration and migrant integration from Austrian election manifestos produced between 1986 and 2013. They remove all party labels, references to previous or subsequent sentences, and gender-sensitive language. They then randomly assign a party cue (i.e, the treatment) to each statement: Green Party (Grüne - Extreme Left), Social Democrats (SPÖ - Center Left), People’s Party (ÖVP - Center Right), or Freedom Party (FPÖ - Extreme Right). The control group is the version without a party cue. An example of a statement without a party cue is ``We stand for a modern and objective immigration policy,'' while a randomly treated statement would be ``We [Greens/Social Democrats/Christian Democrats/Freedomites (`Freiheitliche')] stand for a modern and objective immigration policy.'' Each coder receives 200 statements with randomly assigned conditions, and are subsequently asked whether the statement conveys a positive or negative stance in immigration, or if the statement is neutral or unclear. Finally, \cite{ennser2018impact} test whether the party cue has an effect on the annotation. They find that coders are more likely to rate positively statements with the Greens party cue (i.e., extreme left), and more likely to rate negatively statements with the FPÖ party cue (i.e., extreme right), with no effect from statements with centrist-party cues (e.g., SPÖ and ÖVP). We replicate the results from their experiment in Table \ref{tab:model_original}.  

For our analysis, we use the same statements, and the same instructions given to annotators.\footnote{See the Appendix for an example of the prompt given to the LLMs.} We test two LLM families: OpenAI's ChatGPT (ChatGPT-3.5 Turbo and ChatGPT-4o), and Meta's LLaMa (LLaMa 3-70B and LLaMa 3.1-70B). Each LLM is prompted using the same text that an annotator would have read in \cite{ennser2018impact}'s study. Each prompt is run in a fresh instance of the LLM client, making each response independent from the previous one. Given the independence of each instance, rather than randomly assigning a party cue to each statement (i.e., prompt), we show each LLM all 200 statements with each party cue, including the control. In total, each LLM labels 1000 statements (200 statements x 5 party cues). Since the output from LLMs is stochastic, we increase deterministic answers by setting the temperature to 0 for all runs. Additionally, we replicate each run (200 statements) 10 times, allowing us to measure within model consistency and provide some semblance of replicability. 

To evaluate coding bias, we follow \cite{ennser2018impact} and use an ordered logistic regression to estimate the effect of the treatment on the labelling decisions of the LLMs. To that end, we estimate the following model:

$$y_{ijk} = cue_j + content_i + llm_k + \epsilon_{ijk} $$

\noindent where $y_{ijk}$ is the response of LLM $k$ to statement $i$ with party cue treatment $j$. LLMs categorize each statement as negative, neutral/unclear, positive. Our main variable of interest is $cue_j$, an indicator variable that takes the value of the four party labels (i.e., Greens, SPÖ, ÖVP, and FPÖ), with the \textit{no label} statements as the reference category. We control for content-related factors using either fixed or random effects at the sentence level. We also include fixed effects to capture LLM-specific effects. 

\section{Inter-LLM and Inter-Coder Reliability}

\begin{figure}[!ht]
\centering
\includegraphics[width=10cm]{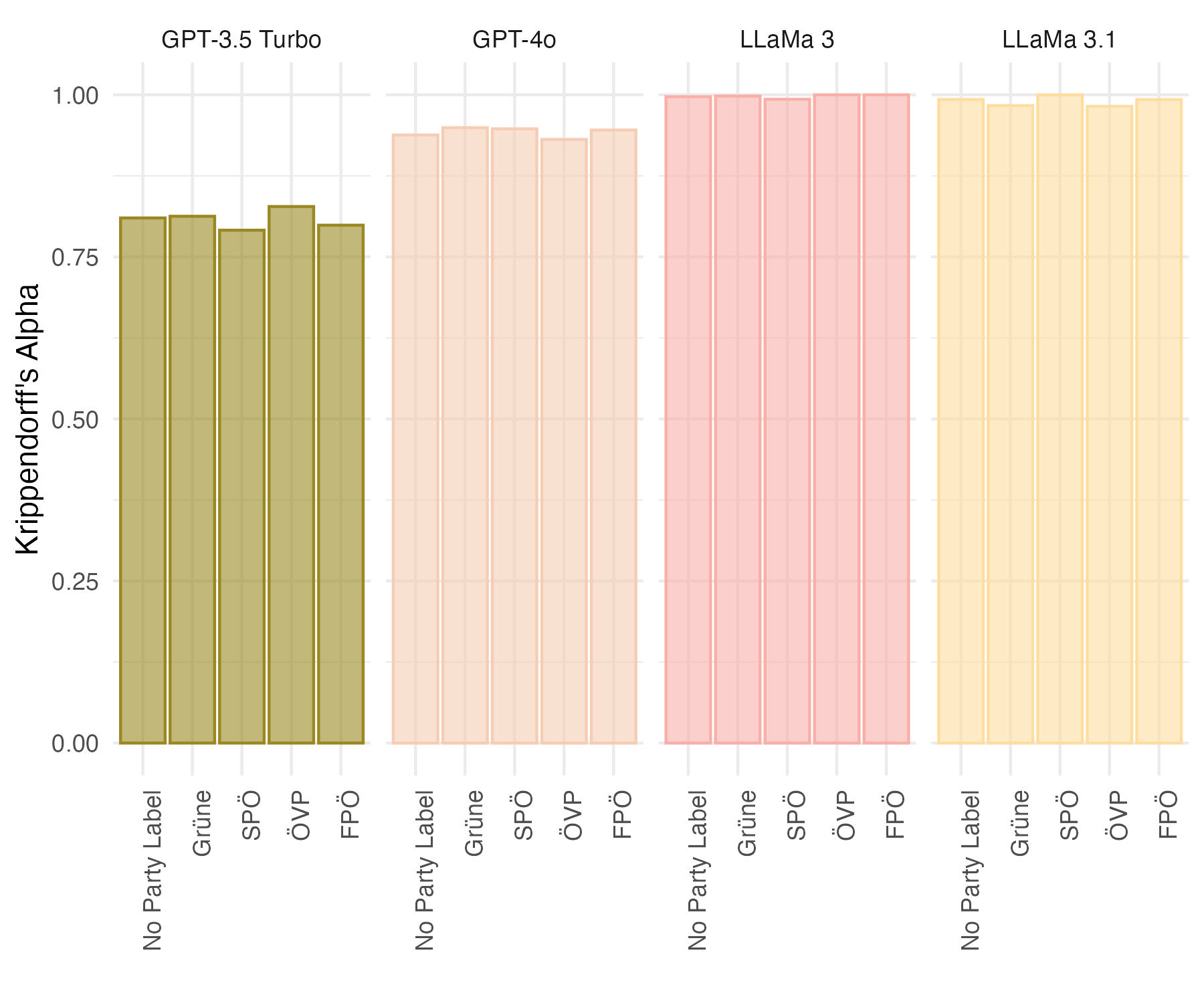}
\caption{Within-LLM consistency.}
\label{fig:within_llm}
\end{figure}

We first describe within-model consistency by looking at inter-run reliability. To this end, we estimate Krippendorff's Alpha, an inter-coder reliability (ICR) score, for each model across the multiple runs (ten in total). Within model Krippendorff's Alpha scores by party cue (treatment) are shown in Figure \ref{fig:within_llm}. Overall, within model consistency is relatively high, with most scores above .8.\footnote{According to \cite{krippendorff2018content}, a Krippendorff's Alpha above .8 is a satisfactory level of agreement, allowing for triangulated inferences based on the labelled data.} The worst performing model is GPT-3.5 Turbo, with an average ICR of .78. There are important differences across LLM families: LLaMa models are highly consistent across runs, with an average Alpha close to 1. Within models, however, there are no important differences across party cues. 

\begin{figure}[!ht]
\centering
\subfloat[Subfigure 1 list of figures text][Within-LLM family ICR]{
\includegraphics[width=7cm]{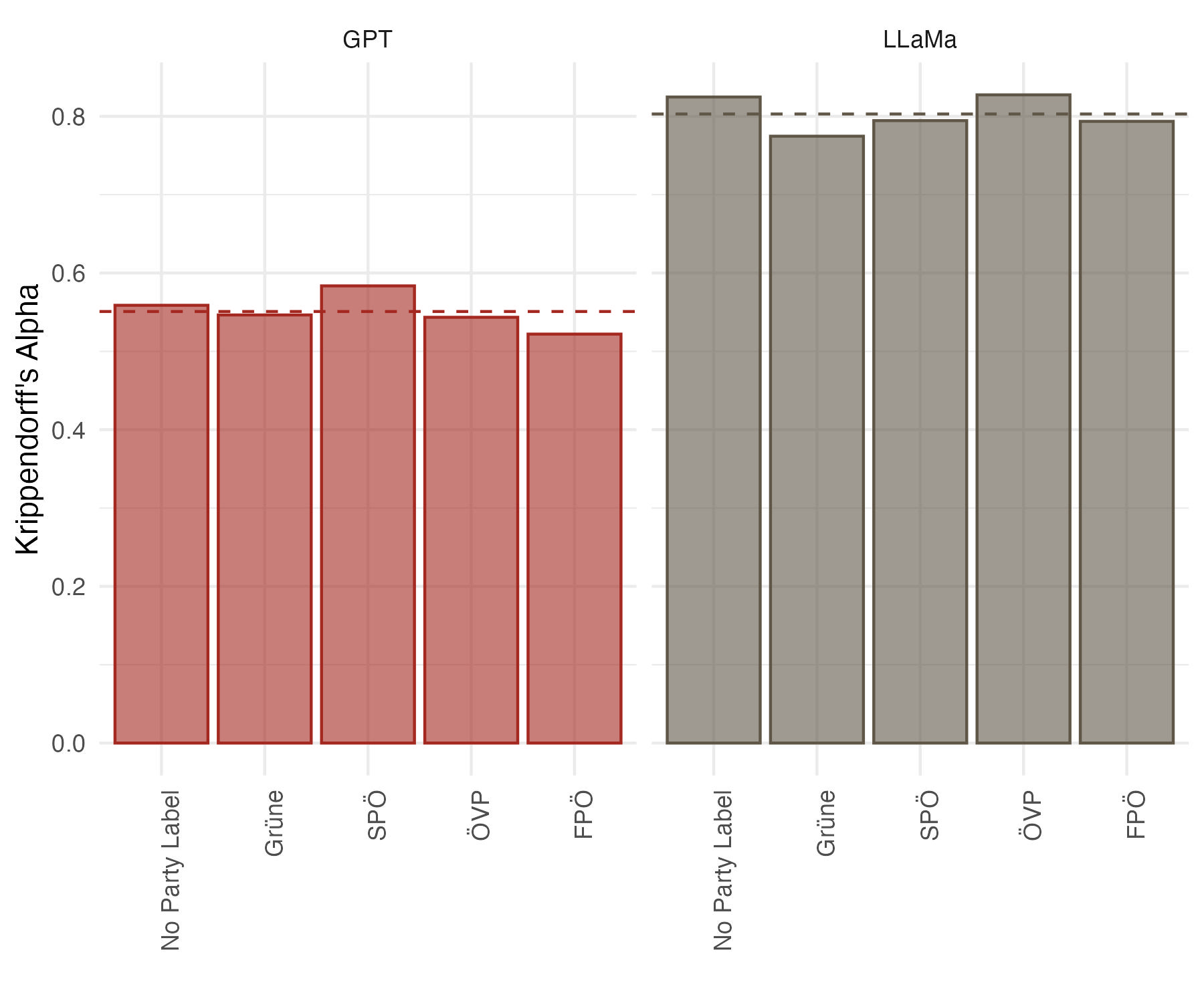}
\label{fig:within_family}}
\qquad
\subfloat[Subfigure 2 list of figures text][ICR across LLMs]{
\includegraphics[width=7cm]{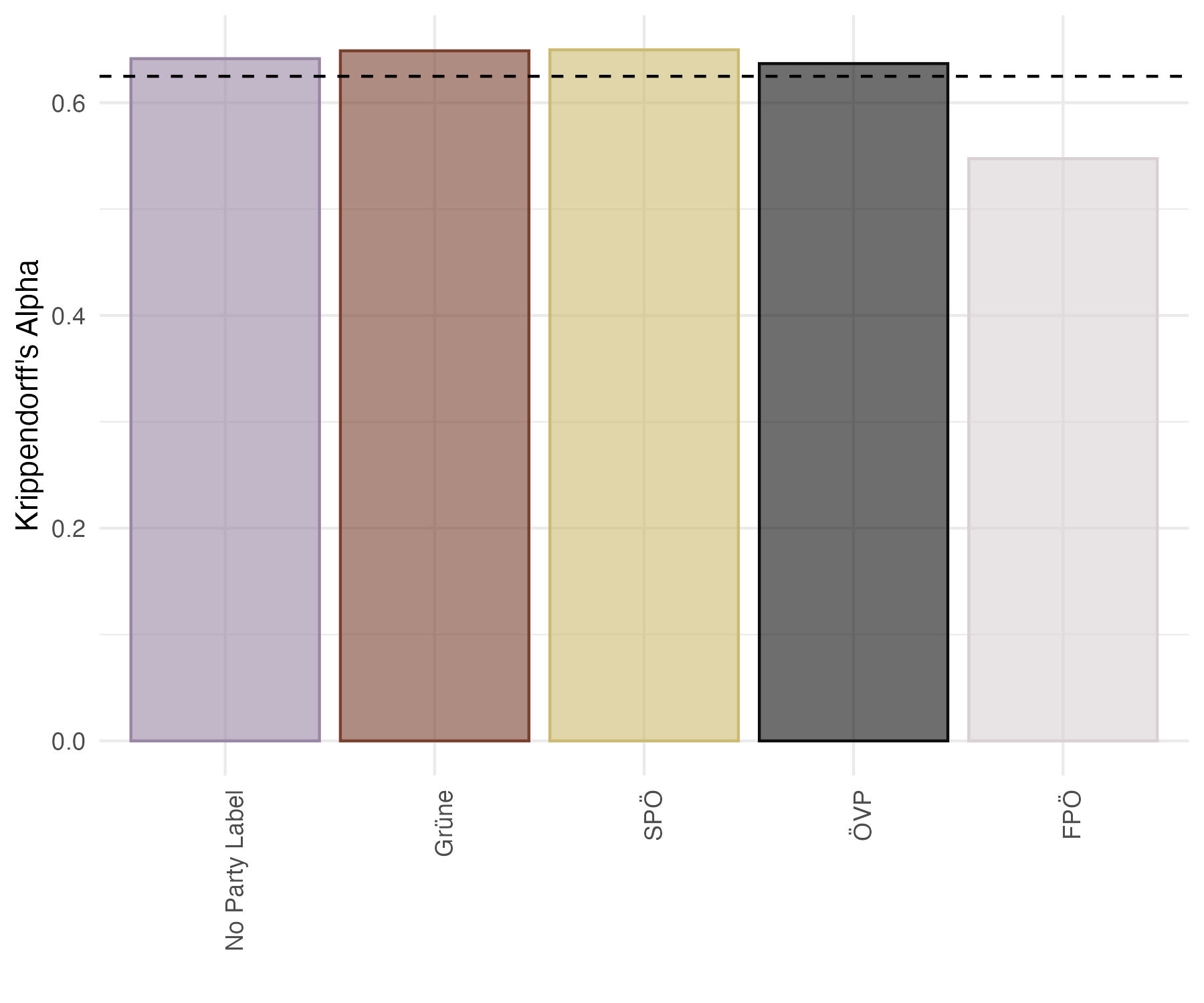}
\label{fig:across_llm}}
\caption{Inter-coder reliability scores within-LLM family, and across all LLMs. ICR scores estimated for each party cue.}
\label{fig:families}
\end{figure}

In Figure \ref{fig:within_family}, we visualize Krippendorff's Alpha for statements with the same treatment between each model family. Since every LLM labels each statement ten times, we adjudicate discrepancies following majority rules.\footnote{In the Appendix we show all results using alternative adjudication methods, but the overall conclusions from the analysis remain unchanged.} The average Alpha score across cues for OpenAI models is .55, considered poor agreement, according to \cite{krippendorff2018content}. We also show Krippendorff's Alpha for inter-LLM agreement when labelling statements with the same party cue (see Figure \ref{fig:across_llm}). Again, the ICR scores are low, averaging around 0.61. The are some important differences between party cues: LLMs agree less about the labels when treated with the FPÖ party cue, than when compared to the rest of treatments. 

\begin{figure}[!t]
\centering
\subfloat[Subfigure 1 list of figures text][ICR using LLMs adjudicated labels]{
\includegraphics[width=7cm]{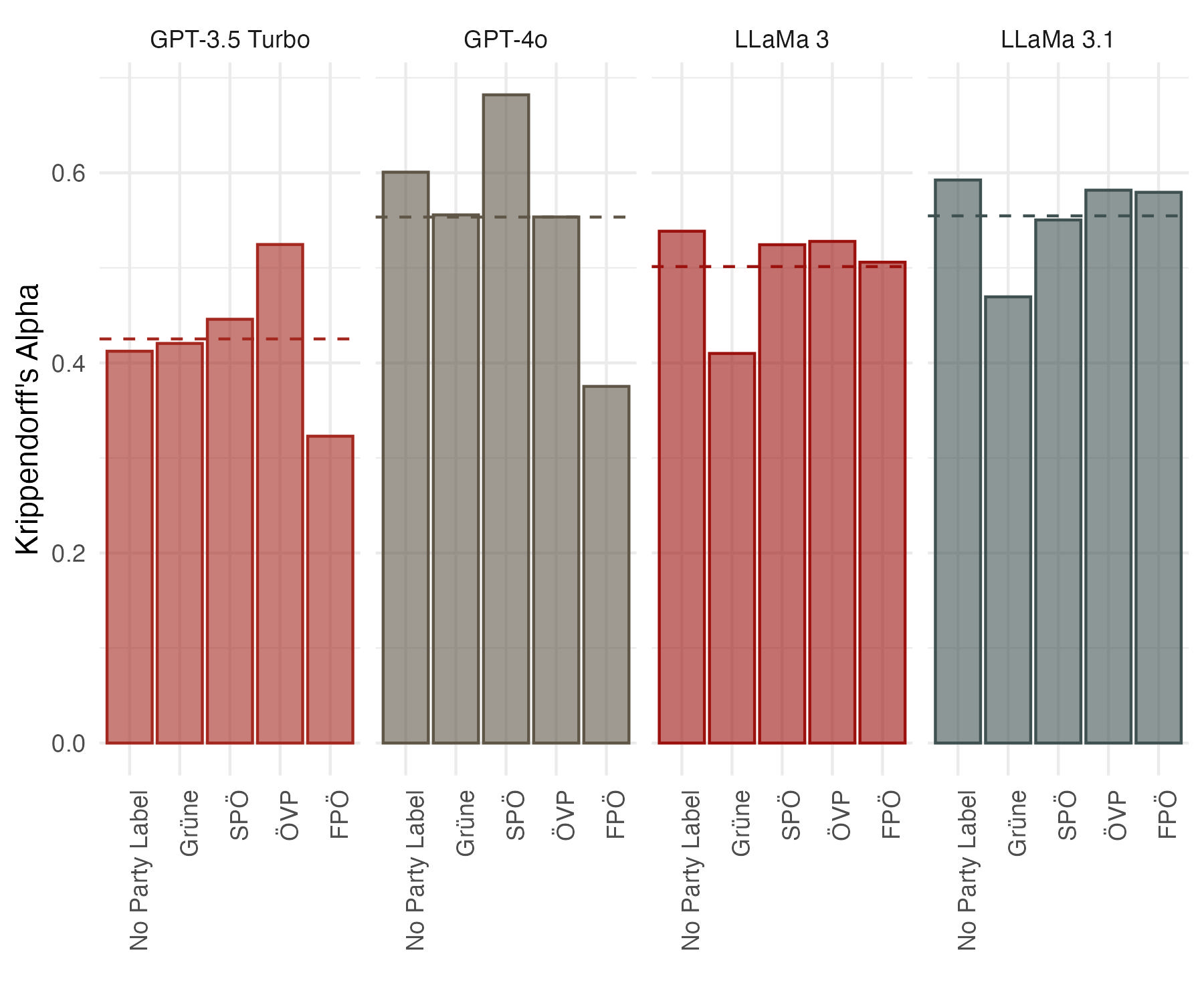}
\label{fig:icr_human}}
\qquad
\subfloat[Subfigure 2 list of figures text][ICR using LLMs labels from all runs]{
\includegraphics[width=7cm]{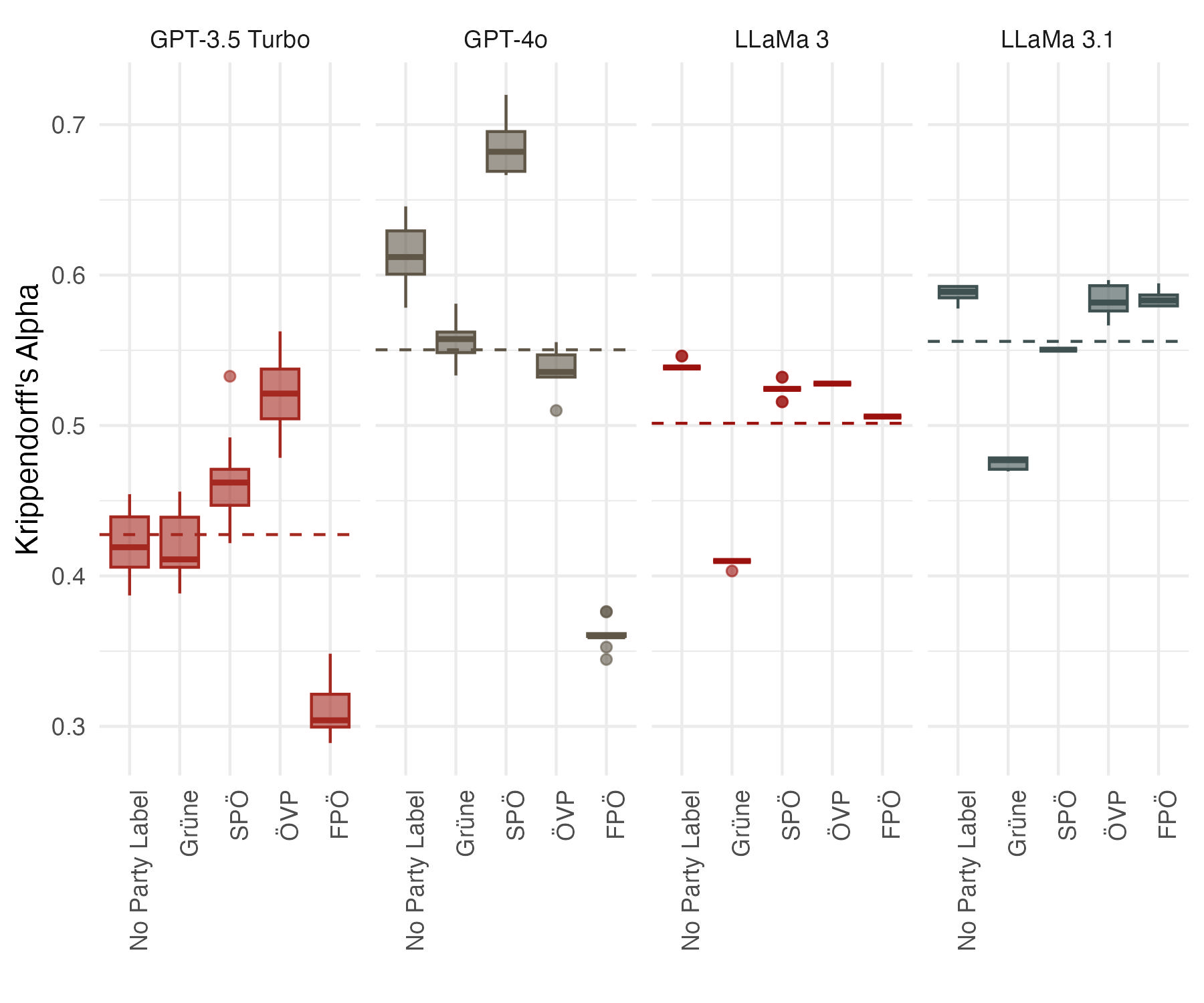}
\label{fig:icr_human_allruns}}
\caption{Inter-coder reliability scores between LLMs and human annotators. On the left, within-model discrepancies across runs are adjudicated using majority rules. On the right, the labels from all runs from every LLM is compared to the labels from human annotators.}
\label{fig:humans}
\end{figure}

Finally, Figure \ref{fig:icr_human} shows ICR scores between LLMs and human annotators from \cite{ennser2018impact}. Similar to the LLM labels used in Figure \ref{fig:within_family}, we adjudicate across-run discrepancies of the same model following majority rules. In terms of ICR, the best performing LLMs are LLaMa 3.1 and ChatGPT-4o, with an average Krippendorff's Alpha of .56 and .55, respectively; the worst performing LLM is ChatGPT-3.5 Turbo with a Krippendorff's Alpha of .43.\footnote{Krippendorff's Alpha below .67 is a poor level of agreement, which does not allow for triangulated inferences from the labelled data. It suggest annotators are not applying the coding scheme consistently \citep{krippendorff2018content}.} The are also differences in the variation of agreement between humans and LLMs across party cues. For the LLaMa family models, the agreement with human coders across party cues is similar. That is not the case for ChatGPT-4o, where there is a Krippendorff's Alpha of 0.68 when labelling statements with the center-left SPÖ label, but 0.37 when labelling statements with the extreme-right FPÖ label. We also estimate Krippendorff's Alpha for each run from every LLM (compared to human coders), and show a boxplot with the distribution of ICR (see Figure \ref{fig:icr_human_allruns}). As expected, given the lower within-model consistency of ChatGPT models, there is greater variation in ICR scores across runs when compared to models from the LLaMa family. Note, however, that there are no overall performance gains from using adjudicated labels, nor labels from a specific run. 

\section{Results}

In Table \ref{tab:model_original}, we replicate the results from \cite{ennser2018impact}, who estimate an ordered logistical regression model where the dependent variable is the label by the coder (e.g., positive, neutral/unclear, negative), and the independent variable of interest if the party cue indicator (using the statement with no party label as the reference category). In their main findings, \cite{ennser2018impact} show that party cues have an effect on coding decisions, but only when the treatment is a party on the ideological extreme (i.e., Greens and FPÖ). They find that coders judge statements on immigration from the Green party more positively, while statements with the FPÖ label are more likely to be labeled negatively (see Table \ref{tab:model_original}). 

\begin{table} [!t] \centering 
\begin{adjustbox}{max width=17cm}
 \begin{threeparttable}
  \caption{Replication of \cite{ennser2018impact}} 
  \label{tab:model_original} 
\begin{tabular}{@{\extracolsep{\fill}}lcc}
\toprule
& Model 1: Fixed Effect & Model 2: Random Effect \\ \midrule 
Grüne           & 1.18***  & 1.06***  \\
& (0.21)   & (0.20)   \\
SPÖ             & 0.00     & 0.00     \\
& (0.21)   & (0.20)   \\
ÖVP             & -0.09    & -0.09    \\
& (0.21)   & (0.20)   \\
FPÖ             & -0.92*** & -0.80*** \\
& (0.21)   & (0.20)   \\
Cut 1: Constant & -1.50*   & -3.15*** \\
& (0.68)   & (0.43)   \\
Cut 2: Constant & 2.05**   & 0.07     \\
& (0.68)   & (0.42)   \\
Num.Obs.        & 2000     & 2000     \\
\midrule 
Statement FE        &  Yes    & No     \\
Statement RE        & No     & Yes     \\
Coder FE        & Yes     & Yes        \\
\bottomrule
\end{tabular}
\begin{tablenotes}
\small
 \item \textit{Note:} Figures are coefficients from ordered logistic regression, with standard errors in parentheses; statement FE (Model 1) not shown. Confidence levels reported as follows: $^{***}$p $<$ .001; $^{**}$p $<$ .01; $^{*}$p $<$ .05. Grüne = Green Party; SPÖ = Social Democrats; ÖVP = People’s Party; FPÖ = Freedom Party.  
 \end{tablenotes}
  \end{threeparttable}
\end{adjustbox}
\end{table}

In Table \ref{tab:model_main}, we estimate a similar ordered logistical regression model where the dependent variable is the label provided by the LLM,\footnote{For Table \ref{tab:model_main}, we use the labels where we adjudicate discrepancies following majority rules. In the Appendix, we estimate the same model using alternative adjudication strategies, as well as all the runs individually, and find similar results.} and the independent variable of interest is the party cue indicator (i.e., treatment). Models 1 and 3 include all aggregated data and control for each LLM, while Models 2 and 4 also include sentence fixed- and random-effect, respectively.\footnote{As a robustness check, in the Appendix we estimate every model only using data of each LLM separately. The results are consistent regardless of model specification.} The positive coefficients from the Green party and SPÖ cues (far-left and center-left) suggest that statements with those cues were judged as more positive by LLMs than statements without party cues. Similarly, LLMs evaluated statements with the ÖVP and FPÖ cues (center-right and far-right) more negatively than statements without party cues. 

The differences in labelling decisions are substantive (as well as statistically significant). When compared to the neutral statements, the Green party (left) cue increases the probability of a statement being coded as `Positive' by 13.9 percentage points, while it decreases the probability that LLMs use the `Neutral' label by 3.0 percentage points, and the `Negative' label by 10.9 percentage points.\footnote{We estimate all predicted probabilities from Model 3, using LLM random-effects, as these yield the more conservative effects.} We observe a similar yet less pronounced effect when applying the SPÖ (center-left) cue: LLMs are 6.7 percentage points more likely to label a statement as `Positive', and 1.0 and 5.6 percentage points less likely to label a statement as `Neutral' and `Negative', respectively. There is an opposite effect for the FPÖ (far-right) cue. For example, LLMs are more likely to identify statements with the FPÖ label as negative--17.2 percentage points more likely--, while it decreases the chances that a statement is coded as `Positive' by 15.9 percentage points. The results are robust to estimations using each LLM sub-sample.\footnote{In the Appendix, we estimate the same model using the data from each individual run, rather than the adjudicated data. The results and conclusions remain unchanged.} The controls provide additional insight into the effect of party cues on labelling decisions. The results from Model 1 in Table \ref{tab:model_main} show that both LLaMa 3 and LLaMa 3.1 are more likely to label statements as 'Positive' than ChatGPT 3.5-Turbo ($p<0.05$).  

\begin{table} [!t] \centering 
\begin{adjustbox}{max width=17cm}
 \begin{threeparttable}
  \caption{The Impact of Party Cues on Labelling Decisions from LLMs} 
  \label{tab:model_main} 
\begin{tabular}{@{\extracolsep{\fill}}lcccc}
\toprule
& Model 1 & Model 2 & Model 3 & Model 4 \\ 
\midrule 
Grüne           & 0.56***  & 2.37***  & 0.56***  & 2.19***  \\
& (0.10)   & (0.18)   & (0.10)   & (0.17)   \\
SPÖ             & 0.27**   & 1.10***  & 0.27**   & 1.03***  \\
& (0.09)   & (0.16)   & (0.09)   & (0.15)   \\
ÖVP             & -0.09    & -0.40**  & -0.09    & -0.37*   \\
& (0.09)   & (0.15)   & (0.09)   & (0.15)   \\
FPÖ             & -0.72*** & -2.93*** & -0.72*** & -2.77*** \\
& (0.09)   & (0.17)   & (0.09)   & (0.17)   \\
ChatGPT-4o      & -0.06    & -0.14    &          &          \\
& (0.08)   & (0.14)   &          &          \\
LLaMa 3         & 0.28**   & 1.17***  &          &          \\
& (0.09)   & (0.14)   &          &          \\
LLaMa 3.1       & 0.33***  & 1.52***  &          &          \\
& (0.08)   & (0.15)   &          &          \\
Cut 1: Constant & -0.60*** & 0.94     & -0.73*** & -2.39*** \\
& (0.09)   & (0.55)   & (0.11)   & (0.55)   \\
Cut 2: Constant & 0.46***  & 5.39***  & 0.33**   & 1.78**   \\
& (0.09)   & (0.57)   & (0.11)   & (0.55)   \\
Num.Obs.        & 4000     & 4000     & 4000     & 4000     \\
\midrule 
Statement FE        &  No    & Yes     &   No   &  No  \\
Statement RE        & No     & No     & No     & Yes     \\
LLM RE        & No     & No     & Yes     & Yes     \\
\bottomrule
\end{tabular}
\begin{tablenotes}
\small
 \item \textit{Note:} Figures are coefficients from ordered logistic regression, with standard errors in parentheses; statement fixed-effects for all fixed-effects models not shown. Confidence levels reported as follows: $^{***}$p $<$ .001; $^{**}$p $<$ .01; $^{*}$p $<$ .05. Grüne = Green Party; SPÖ = Social Democrats; ÖVP = People’s Party; FPÖ = Freedom Party.  
 \end{tablenotes}
  \end{threeparttable}
\end{adjustbox}
\end{table}

There are two important elements to note from comparing the effects of party cues on human and LLM coders. First, as \cite{ennser2018impact} suggest, coders use prior knowledge to contextualize information provided by text. For humans, prior knowledge comes from heuristic processing; for LLMs, from the context in which party labels appear in pre-training data (i.e., the massive corpora used to train LLMs). Both human and LLMs coders seem to understand contextual cues in the same way: the direction of the bias for left-leaning parties and right-leaning parties match with prior expectations about their position vis-a-vis immigration (i.e., left-leaning parties are more likely to frame immigration in a positive light; right-leaning parties are more likely to frame immigration in a negative light). Second, for LLMs party cues appear to have greater weight on the decision to label statements one way or another. For example, unlike with human coders, the center-left SPÖ and the center-right ÖVP party labels have a significant effect on labelling decisions. In the Appendix we show that, when estimating the models using all the runs, the effect of cues is also statistically significant for all parties. Furthermore, the magnitude of the effect of all party cues is greater in LLMs than in human coders. The biggest effect of party cues on human coders come from the Green party treatment. Compared to the no party label, the Green party label increases the probability of a statement being coded as `Positive' by 22 percentage points, while decreasing the probability that human coders use the `Neutral' and `Negative' categories by 19 and 3 percentage points, respectively \citep{ennser2018impact}. When using the same model specification for LLMs (coder fixed-effect and sentence random-effects in Model 4), the Green party cue increases the probability of a statement being coded as `Positive' by 45.61 percentage points, while decreasing the probability that LLMs use the `Neutral' and `Negative' categories by 38.3 and 7.39 percentage points, respectively.

\section{Conclusion/Discussion}

Our results show that, similar to human coders, contextual information affects responses from LLMs. In particular, we provide evidence of how party cues effect the way that LLMs label policy statements: left and center-left party cues increase the probability an immigration related statement will be labelled as positive; right and center-right party cues increase the probability an immigration related statement will be labelled as negative. Unlike humans, this effect is not limited to parties at the ideological extremes. We also find that the magnitude of the effect is greater for LLMs than it is for human coders. 

\cite{ennser2018impact} argue that human coders use prior knowledge to contextualize information, what they call heuristic processing, and that this contextualization will have an effect on policy-related labelling decisions. We argue for a similar intuition when it comes to LLMs. LLMs are trained using troves of human-generated text data, and high valence tokens, such as party labels, are more likely to appear under specific contexts. Since the data reflects biases from society, these biases will also be picked up by LLMs. Given the obscurity of how LLMs are trained, it is difficult to assess the degree to which our proposed mechanism is actually affecting the decisions taken by LLMs. However, future research can expand on the types of cues, and types of tasks, that affect the behavior of LLMs.

The context-based decision-making observed from LLMs is not necessarily problematic. Informational cues can improve the validity of the data as long as the priors of LLMs are 'correct.' However, the impossibility of realistically perusing the data used to train LLMs, or of understanding the `thought process' behind any given response, makes it unlikely for researchers to know what these priors are, and which tokens (or set of tokens) have these priors embedded.\footnote{Future research should explore how fine-tuning LLMs--using training data to customize the behavior of LLMs to a specific task--can adjust the priors of LLMs to provide more accurate responses.} More importantly, with the constant updating of LLMs, as well as the proliferation of new LLMs, it is difficult to know whether priors will all be the same across platforms and time. 

Our findings also speak more broadly to the behavior of LLMs. In this research note, we use 10 runs to account for the probabilistic nature of text generation \citep{c2024bert}. We find that internal consistency (i.e., the consistency of results across multiple runs) varies greatly across LLM families and LLMs. This has important implications, not only for the use of LLMs as coders, but in the replicability of outputs from LLMs in research. It also has important implications on the adjudication choices made by researchers. 

Overall, our research cautions researchers who are considering using LLMs as coders, echoing previous research that has highlighted biases from human coders \citep{benoit2016crowd,ennser2018impact,laver2000estimating}. While there are many benefits to using LLMs as annotators, such as their low cost and high accuracy \citep{gilardi2023chatgpt,heseltine2024large}, as with any other labelling task, proper validation is paramount. 
 
\pagebreak

\vspace{1cm}
\setstretch{1}
\bibliography{reference.bib}
\bibliographystyle{apsr}

\end{document}